# Mapeo de controversias a partir de inteligencia artificial: análisis sobre el conflicto Hamás-Israel en YouTube

# Mapping Controversies Using Artificial Intelligence: An Analysis of the Hamas-Israel Conflict on YouTube

**Autores:**

**Víctor Manuel Hernández López**

vimaherlo123@gmail.com | vhernandez56@areandina.edu.co

**Jaime E. Cuellar**

jaimecuellar@javeriana.edu.co

## Resumen

Este artículo analiza la controversia Hamás-Israel a través de 253,925 comentarios en español de YouTube, publicados entre octubre de 2023 y enero de 2024, tras el ataque del 7 de octubre que intensificó el conflicto. Utilizando una aproximación interdisciplinaria, el estudio combina el análisis de controversias desde los Estudios Sociales de Ciencia y Tecnología (CTS) con metodologías computacionales avanzadas, específicamente el Procesamiento de Lenguaje Natural (PLN) con el modelo BERT (Bidirectional Encoder Representations from Transformers). A través de este enfoque, los comentarios fueron clasificados automáticamente en siete categorías, reflejando posturas pro-palestinas, pro-israelíes, anti-palestinas, anti-israelíes, entre otras. Los resultados muestran un predominio de comentarios pro-palestinos, aunque los comentarios pro-israelíes y anti-palestinos reciben más "likes". Este estudio también aplica la teoría de agenda setting para evidenciar cómo la cobertura mediática influye en la percepción pública, observando un cambio significativo en la opinión pública, que transitó de una postura pro-palestina a una más crítica hacia Israel. Este trabajo destaca la importancia de combinar perspectivas de ciencias sociales y herramientas tecnológicas en el análisis de controversias, y presenta una innovación metodológica al integrar análisis computacionales con teorías sociales críticas para abordar complejos fenómenos de opinión pública y narrativa mediática.

## Abstract

This article analyzes the Hamas-Israel controversy through 253,925 Spanish-language YouTube comments posted between October 2023 and January 2024, following the October 7 attack that escalated the conflict. Adopting an interdisciplinary approach, the study combines the analysis of controversies from Science and Technology Studies (STS) with advanced computational methodologies, specifically Natural Language Processing (NLP) using the BERT (Bidirectional Encoder Representations from Transformers) model. Using this approach, the comments were automatically classified into seven categories, reflecting pro-Palestinian, pro-Israeli, anti-Palestinian, anti-Israeli positions, among others. The results show a predominance of pro-Palestinian comments, although pro-Israeli and anti-Palestinian comments received more "likes." This study also applies the agenda-setting theory to demonstrate how media coverage significantly influences public perception, observing a notable shift in public opinion, transitioning from a pro-Palestinian stance to a more critical position towards Israel. This work highlights the importance of combining social science perspectives with technological tools in the analysis of controversies, presenting a methodological innovation by integrating computational analysis with critical social theories to address complex public opinion phenomena and media narratives.

## Palabras clave:

Controversia Hamás-Israel, YouTube, inteligencia artificial, BERT, Agenda Setting

**Introducción**

El 7 de octubre de 2023, la noticia de que Hamas había lanzado misiles desde la Franja de Gaza a un grupo de ciudadanos israelíes avivó la guerra con Israel (Qué es Hamás, el grupo islamista militante que lanzó un ataque sin precedentes contra Israel, 2023). Israel respondió bombardeando zonas de la Franja misma, marcando el inicio del escalamiento del conflicto alrededor de las zonas israelíes y palestinas (Guerra entre Israel y Gaza, 8 de octubre - Más de 1.100 muertos en el ataque de Hamás y la represalia de Israel, 2023). Lo cierto es que, aunque recrudecido, el conflicto tiene más de medio siglo de historia y se origina luego de que actores israelíes de carácter sionista se establecieron en el territorio que antes de la Segunda Guerra Mundial se denominaba Palestina y fundaron el Estado de Israel.

El sionismo nace como un movimiento político que enfrentaría al antisemitismo europeo en la segunda parte del siglo XIX (Katz, 1979). Este movimiento recalca la idea de emancipación del pueblo judío y la "concesión de un estado moderno" como lo han obtenido los pueblos cristianos e islámicos (Katz, 1979). Sin embargo, a pesar de sus ideales fundacionales, autores señalan al movimiento sionista de encabezar cruzadas con la intención de "reconstruir Palestina y reestablecer Israel en 1948". Se comprende entonces que el proyecto israelí tomó una idea de unión social para realizar un "proyecto de transformación [...] que racionalizó la erradicación de la realidad en Palestina" (Said, 1979, p. 9).

Es precisamente en este contexto en el que décadas después nace Hamás, que se autodenomina como un "movimiento de resistencia" y frente político islamista para enfrentar la ocupación israelí alrededor del año 1987 (Abu-Amr, 1993). En los años siguientes, Hamás se negó a hacer parte de organismos como la Autoridad Nacional Palestina y adelantó atentados suicidas, ataques y otros hechos violentos que le han valido, hasta nuestros días, ser reconocido como grupo terrorista por múltiples gobiernos. En el 2006, el grupo se presentó a las elecciones legislativas y obtuvo el control de la Franja de Gaza, que gobierna desde entonces ofreciendo servicios sociales además de su vocación política. Más de dos millones de personas habitaban la franja antes de los atentados del 7 de octubre de 2023, en medio de la crisis humanitaria ocasionada por el bloqueo económico de Israel al territorio.

A partir de los acontecimientos de octubre de 2023, numerosos expertos, organizaciones y medios de comunicación han seguido de cerca los hechos. Lo mismo ha ocurrido en redes sociales, donde personas en todo el mundo han expresado sus opiniones sobre el conflicto. Por una parte, algunos argumentan la justa insurgencia por parte de grupos armados palestinos que consideran ilegítimo el poder de Israel y lo señalan de ser un "*Estado colonialista*" y "*usurpador de tierras palestinas*".

Por otro lado, existen personas que mencionan que *"Israel tiene derecho a defender su territorio"*, justificando la ofensiva a partir del lanzamiento de misiles de Hamás y calificando los bombardeos israelíes subsecuentes como parte de la dinámica de la guerra. La naturaleza actual, polarizante y diversa de estas posiciones hace manifiesta la existencia de una controversia desde la perspectiva de Zielinski et al. (2018).

Existen múltiples metodologías que permiten analizar controversias. Este artículo combina el mapeo de controversias realizado por los Estudios Sociales de Ciencia y Tecnología (CTS) y el análisis de controversias realizado desde las ciencias computacionales y estadísticas en una apuesta novedosa que busca conjugar las fortalezas de ambos. Para este fin, se analizarán comentarios en español realizados en vídeos de medios informativos en YouTube. Nace así la pregunta de investigación: ¿cuál es el mapa general de la controversia Hamas-Israel que resulta del análisis de los comentarios en español de vídeos de YouTube posteados entre octubre de 2023 y enero de 2024?

Para analizar esto, el presente artículo se divide en tres secciones: la primera, una revisión de literatura de los conceptos de mapeo de controversias desde los dos enfoques empleados y del conflicto Israel-Palestina desde las redes sociales; la segunda, la metodología mixta que se emplea para un uso crítico de las herramientas de inteligencia artificial junto con el *agenda setting,* que reflexiona sobre cómo los medios de comunicación influyen en la percepción pública y en la búsqueda de información en situaciones de incertidumbre; y finalmente, una tercera que da cuenta del mapeo de la controversia y la analiza críticamente.

**Revisión de literatura**

**Mapeo de controversias**

Zielinski et al. (2018) definen controversia como:

> Discusión pública sobre un asunto que tiene partidarios en un lado y mucha gente que discrepa fuertemente de ella o se sorprende de alguna manera por esta. Con respecto al tiempo, la controversia se caracteriza por el hecho de que no se llega a un acuerdo o consenso sobre el tema durante un período prolongado. Las controversias van desde fuertes polémicas hasta discusiones educadas y ordenadas (pág. 16).

En este artículo se entiende por controversia como una discusión de carácter público con posiciones diversas de las cuales las personas toman parte. En los CTS, el mapeo o cartografía de controversias se ha usado históricamente para revisar las distintas opiniones que ha tenido la ciencia respecto a un tema en específico (Whatmore, 2009). Los CTS han tomado el mapeo de controversias como una metodología propia, puesto que fue propuesta originalmente por Bruno Latour y posteriormente reformulada por autores como Venturini y Munk (2021). Latour, precursor de esta aproximación, buscaba llevar a los principios de la Teoría del Actor-Red (TAR) a la práctica (Venturini, 2009).

Por su parte, las ciencias computacionales entienden el análisis de controversias más como como método computacional que "implica el uso de técnicas computacionales para detectar, analizar y visualizar la comparación pública sobre asuntos de actualidad" (Marres, 2015, pág. 3). Desde esta perspectiva es común explorar controversias limitándose a la automatización de procesos de clasificación y graficación de redes (Németh, 2023; Shiri et al., 2013; Coletto, 2017; Myungha et al., 2016; Moats et al., 2019). En este artículo se utilizaron métodos computacionales para el análisis detallado de controversias. Se entiende entonces que este artículo pretende mediar entre los CTS y la ciencia de datos a partir de un uso de métodos computacionales para la descripción de la controversia misma, ofreciendo herramientas útiles para la comunicación y las ciencias sociales.

A continuación se presentan los principales antecedentes asociados a ambos enfoques. Es de notar que los análisis de controversias ricos en discusión, especialmente asociados a controversias políticas, provienen de las ciencias sociales (Németh, 2023). De la misma manera, se presentan investigaciones en controversias que buscan entender sus aspectos metodológicos o técnicos, como se profundiza en el siguiente apartado.

**Desde la ciencia de datos**

Por parte de la ciencia de datos Dori-Hacohen (2013) y Jang (2016) exploran el uso del aprendizaje automático para detectar controversias en la web, con el enfoque de clasificación k-vecino más cercano de Dori-Hacohen que demuestra una mejora del 22% sobre un sentimiento-método basado, y el marco probabilístico de Jang lograr una mejora relativa significativa en AUC. Habernal (2015) extiende este trabajo aprovechando los portales de debate para la minería de argumentación semi-supervisada, lo que demuestra un aumento significativo del rendimiento en la identificación de los componentes del argumento. Estos estudios colectivamente destacan el potencial del aprendizaje automático para avanzar en el mapeo de controversias.

Es precisamente Németh (2023) quien ubica en su estado del arte el Procesamiento de Lenguaje Natural (PLN) con un potencial muy alto para la investigación de la polarización política. Además, usar PLN "también proporciona argumentos para la integración de paradigmas explicativos y predictivos, y para un enfoque más interdisciplinario de la investigación sobre la polarización" (pág. 308). Algunas investigaciones realizadas utilizan *word embedding* para la identificación de *features* que se usarán en modelos de clasificación. Otros artículos utilizan *topic modeling* y análisis de sentimientos para analizar controversias, en su mayoría, de carácter políticos-electorales (Németh, 2023).

Por su lado, Godara y Kumar (2019) estudian cinco algoritmos de aprendizaje automático (K-Vecinos Más Cercanos (KNN), Árbol de Decisión, Redes neuronales artificiales (ANNs), Bayes Naïve y Support Vector Machine (SVM)) y concluyen que SVM es el más usado en clasificación por su desempeño en la clasificación (performance). Por su parte, De Zarate et al. (2020) comentan que los modelos FastText y Bidirectional Encoder Representations from Transformers (BERT) tienen un muy buen desempeño a la hora de detectar controversias. Los autores precisan que al ser BERT una red neuronal es mucho más poderosa su detección de controversias y posiblemente pueda presentar falsos positivos. De igual forma, Garimella et al. (2016) recalcan que los cuatro análisis más utilizados en el mapeo de controversias desde las ciencias computacionales son:

> (i) estructura de las aprobaciones, es decir, quién está de acuerdo con quién sobre el tema, (ii) estructura de la red social, es decir, que está conectado con quién entre los participantes en la conversación, (iii) contenido, es decir, las palabras clave utilizadas en el tema, (iv) sentimiento, es decir, el tono (positivo o negativo) utilizado para discutir el tema (pág. 34).

Así, análisis recientes que exploran controversias a partir de procesos computacionales muestran una gama amplia de modelos posibles a utilizar que se centran en clasificación, estructura y análisis de sentimientos.

**Desde las ciencias sociales**

Por parte de las ciencias sociales, investigaciones recientes han logrado avances significativos en el campo del mapeo de controversias, particularmente en el contexto de los medios digitales. Por ejemplo, Marres (2015) enfatiza la necesidad de abordar el sesgo digital en el análisis de controversias, proponiendo un cambio del análisis de controversias al mapeo de temas, el cual propone utilizar métodos computacionales en el análisis de una controversia teniendo en cuenta sus limitaciones y sesgos. Algunos asuntos importantes a los que invita Marres (2015) cuando se trabaja con métodos digitales es a cuidar el proceso de etiquetado.

Además de identificar la controversia y contextualizarla en tiempo y características sociales, el análisis de controversias utiliza herramientas computacionales para realizar el respectivo análisis. Venturini y Munk (2022) ubican de igual forma el mapeo de controversia como una metodología propia de la TAR propuesta por el filósofo y sociólogo Bruno Latour. Aunque el método se sitúa en la investigación de los debates sociotécnicos y en general en estudios sociales de la ciencia, cada vez más se utiliza para mapear cualquier tipo de controversia (Marres, 2015). Para Venturini y Munk (2022), los elementos metodológicos fundamentales de un mapeo de controversia son: seguir los actores, ver las proporcionalidades y los pesos de la controversia, no olvidar nuestra posición dentro de la controversia, estudiar los actores dentro de sus contextos propios, realizar visualizaciones legibles y dejar los datos abiertos para futuras investigaciones (pág. 7).

De esta manera, el presente artículo intenta mediar entre lo analítico de un mapeo realizado desde CTS y la utilización de modelos predictivos desde la ciencia de datos, para generar una aporte metodológico holístico que sirva en los campos de la comunicación y las ciencias sociales en general.

**Conflicto Israel-Palestina en redes sociales**

Existen varios antecedentes que han analizado el conflicto en redes sociales que incluyen Reddit, X y Youtube. Dado que en la mayoría de ellos se habla del conflicto en clave de Israel-Palestina, no de Israel-Hamás, así se titula esta sección. Nushin et al. (2024) realizan un estudio sobre el conflicto Israel-Palestina en Reddit, utilizando herramientas computacionales como Vader y modelos de aprendizaje automático para el análisis de sentimientos. Además, emplean el modelo de Latent Dirichlet Allocation (LDA) para realizar *topic modeling* con el fin de analizar las perspectivas geopolíticas de los usuarios. La principal conclusión de este estudio es que la mayoría de las personas mantienen una postura neutral respecto al conflicto. Por su parte, Guerra et al. (2024) también analizan Reddit, pero con un enfoque en las opiniones extremas sobre el conflicto, utilizando un enfoque de lexicón usando VADER y TextBlob. Identifican que los eventos que generaron las opiniones más extremas fueron los bombardeos de las Fuerzas de Defensa de Israel contra el Hospital Al Quds y el Campo de Refugiados de Jabalia, así como el fin del alto el fuego tras un ataque terrorista.

Por un lado, Qiu et al. (2024) analizan comentarios en X sobre el conflicto Israel-Palestina utilizando LDA para *topic modeling*, el NRC Emotion Lexicon para análisis de sentimientos y modelos no supervisados como NLI con enfoque "zero-shot" (es decir, sin entrenamiento adicional). Teóricamente, se basan en la Moral Foundations Theory (MFT). Los resultados indican que en esta red social predominan preocupaciones relacionadas con el daño, la degradación y la traición. Se concluye que tanto los tuits anti-Hamás como los anti-Israel se centran en los conflictos, las guerras y los derechos humanos, con una fuerte carga emocional negativa. Los primeros reflejan una mayor intensidad emocional, mientras que los segundos abordan una gama más amplia de temas. Alamsyah et al. (2024) examinan el conflicto de Gaza a través de la red social X, realizando un análisis de redes enfocado en la modularidad para identificar clústeres y evaluar la polarización. Los resultados evidencian una fuerte polarización, donde el 82,58 % de los nodos corresponde a posiciones pro-Palestina y el 17,42 % a posturas pro-Israel, con una mayor interacción dentro de cada grupo que entre grupos opuestos.

Por otro lado, Rico-Sulayes (2025) investiga la detección automática de lenguaje de odio en el contexto del conflicto Israel-Palestina a partir de comentarios en redes sociales como YouTube y X. Su objetivo es "producir un corpus que permita entrenar un algoritmo para detectar el lenguaje de odio y apoyo característico del conflicto Israel-Palestina" (p. 13). Para ello, establece cuatro categorías que inspiran las planteadas en este documento: dos categorías pro-Palestina (1. odia hacia Israel y 2. apoyo hacia Palestina) y dos pro-Israel (3. odio hacia Palestina y 4. apoyo a Israel). De manera similar, Maathuis y Kerkhof (2024) proponen el uso de Grandes Modelos de Lenguaje (LLM) para categorizar e interpretar los debates sobre el conflicto Israel-Palestina, utilizando datos extraídos de YouTube. Como resultado, identifican diversas narrativas y clústeres en el corpus analizado. Asimismo, Liyih et al. (2024) emplean técnicas de aprendizaje profundo para realizar un análisis de sentimientos de polaridad (positivo, negativo o neutral) en comentarios de YouTube relacionados con el conflicto Israel-Palestina. Esto da cuenta de la relevancia que ha adquirido los comentarios de YouTube como fuente de datos para analizar fenómenos sociales como el conflicto antes mencionado.

**Metodología**

Esta investigación sigue un análisis mixto de triangulación (Creswell y Creswell, 2018) que trabaja de manera simultánea con lo cuantitativo y lo cualitativo. El componente cuantitativo se desarrolló a través de PLN. Por su parte, el cualitativo se hizo a partir del análisis crítico y contextual de los comentarios, que sirvió para entrenar el modelo, al tiempo que se empleó un análisis de *agenda setting* que proponen McCombs y Valenzuela (2007) como la apropiación pública de las agendas de los medios de comunicación.

Para especificar estas ideas y teniendo en cuenta lo propuesto por Venturini (2010) y Venturini y Munk (2022), la metodología de este artículo tiene en cuenta los siguientes puntos:

1. Se realizó una recolección de comentarios escritos en español en vídeos de canales de YouTube de prensa que resultan de la búsqueda Hamás - Israel - Gaza - Palestina y que han sido publicados desde el 7 de octubre del año 2023 al 7 de enero de 2024. La elección de esta temporalidad es relevante porque da cuenta de los meses iniciales de la controversia reciente y sus reacciones. Es importante recalcar que, al ser una investigación que hace uso de PLN, los datos deben estar contextualizados, correctamente recolectados y dispuestos para análisis.
2. Se llevó a cabo un preprocesamiento de la base de datos utilizada en el modelo. Esto incluye la limpieza de comentarios vacíos, comentarios con letras o emojis sin referencia a la controversia, además de comentarios realizados por fuera de las fechas propias del periodo elegido.
3. Se llevó a cabo un proceso de anotación basado en caracteres, utilizando los comentarios seleccionados aleatoriamente. Esta escogencia aleatoria garantiza que los datos no sean de un mismo vídeo y que no presenten sesgos de agrupamiento. El análisis de los comentarios aleatorios se realizó por los investigadores para la selección de clasificaciones de la controversia. Esta anotación fue llevada a cabo de la forma "MATTER+PD" propuesta por Finlayson y Erjavec (2017).
4. Se utilizó el modelo BERT para la clasificación de texto. Todo el código fue realizado en Python versión 3.
5. Se establecieron las discusiones generadas de los comentarios dentro de las clasificaciones generadas y dispuestas por cantidad, temporalidad y afinidad o "me gustas".

**Proceso detallado:**

Se enlistó un total de 289 vídeos de YouTube en español resultado de las búsquedas: "Hamas español", "Israel español", "Gaza español" y "Palestina español". Esto con el fin de atender a vídeos que tuviesen los principales actores de la controversia y asegurar el lenguaje de los mismos.

Los vídeos, en su mayoría, provienen de páginas de noticias internacionales como DW, CNN, Euronews, entre otros. En pequeña proporción aparecen vídeos de creadores de contenido sin afiliación periodística y vídeos que pretenden educar y explicar contextos desde y para comunidad general. La captura de los comentarios se realizó con YouTube Data API v3 el 8 de enero de 2024. La base de datos generada contiene 270.573 filas y 6 columnas en las que se incluye: autor, fecha de publicación, número de likes, texto del comentario, ID del vídeo, y estatus (si es público o no).

Una vez obtenidos los datos, se procedió a la ideación de las posibles categorías existentes. Para ello se tomó una base de subdatos de 500 observaciones elegidas aleatoriamente dentro de la base de datos total. En esta investigación, el proceso de la elección de las etiquetas a utilizar surgió de una identificación no automatizada de entidades y disposiciones textuales dentro de cada uno de los comentarios revisados. Las etiquetas enlistadas son el resultado entonces de un preanálisis de los comentarios que resultan en tipos de clasificaciones sugeridas explícitamente por los mismos datos y escogidos por los investigadores, además de la revisión de literatura (Rico-Sulayes, 2025). Este proceso se denomina "anotación basada en caracteres", según Puestejovsky y Stubbs (2012). Así, la inclusión o exclusión de un comentario dentro de una etiqueta u otra está basada específicamente en su sintaxis gramatical y no tanto en su significado implícito. Lo anterior con el fin de tomar decisiones de clasificación lo suficientemente parecidas a las que la máquina podría llegar a realizar. Al final, cada una de las etiquetas dispone de un valor numérico asignado para mejor rendimiento del modelo a desarrollar.

A continuación, se especificará en qué consiste cada etiqueta:

1. *Comentarios no relacionados (NR - Label: 4):* aquellos comentarios que no tienen relación con la controversia analizada. Esta categoría engloba comentarios de algunos vídeos donde no solamente presentaban noticias relacionadas con el conflicto Hamás-Israel, sino también resúmenes de otras noticias internacionales. Así, se presentaban temas no relacionados con la controversia analizada. De la misma manera, algunos comentarios atacaban a personas entrevistadas por sus condiciones políticas sin conexión explícita con la controversia.

2. *Comentarios sin postura (SP - Label: 3):* comentarios que no toman postura explícita hacia Israel o Hamás, o que atacan a ambos bandos por igual. Aquí se ubican aquellos comentarios que atacaban explícitamente el conflicto en general diciendo *"Paz para el mundo"* o *"La guerra no es el camino"*. Igualmente, en esta categoría se encuentran los comentarios que atacan explícitamente a ambos "lados" del conflicto cuando mencionan que *"Tanto Israel como Hamás deben pagar por lo que hacen".* Esta categoría no entra en el análisis de controversia porque no toma una postura definida hacia alguno de los dos actores. Muchos de los comentarios de etiqueta SP alegorizan a Dios como único poder capaz de acabar con la guerra. En esta etiqueta también se ubican comentarios a noticieros de los vídeos subidos o presentadores de los mismos, gobiernos alrededor del mundo, ideologías económicas y políticas, y otros elementos que no tenían que ver directa o explícitamente con la controversia analizada.

3. *Comentarios en apoyo al pueblo palestino (Pro-palestino - Label: 6):* comentarios que apoyan al pueblo palestino, por ejemplo calificando la ofensiva como genocidio. Aquí se engloban aquellos comentarios que escriben *"Palestina libre"* o que se muestran explícitamente preocupados por la situación de la población que habita la Franja de Gaza. Algunos ejemplos de esta etiqueta son: *"Viva Palestina!", "PAZ, JUSTICIA Y EQUIDAD también para PALESTINA…".*

4. *Comentarios apoyo al Estado de Israel (Pro-Israel - Label: 5):* comentarios que apoyan a Israel y su defensa de Hamás. Esta categoría acoge comentarios que explícitamente defienden el actuar de Israel, comentan *"Dios bendiga a Israel"* o se muestran preocupados por la situación que vive la población israelí. De igual manera señalan a Hamás por atacar primero y defienden la legítima defensa del pueblo de Israel. Aquí también se encuentran los comentarios más explícitamente violentos por el uso de "groserías" o palabras soeces: *"Estan atacando a Israel y los HP de Iran no quieren quedae defienda…"[SIC].*

5. *Comentarios que señalan a Israel (Anti-Israel - Label: 1):* comentarios que señalan a Israel de cometer actos inmorales y lo consideran el principal culpable del conflicto. Algunos comentarios ubican a Benjamin Netanyahu, primer ministro de Israel, al gobierno y políticos sionistas como principales responsables. De la misma manera esta etiqueta engloba aquellos comentarios que señalan explícitamente a Israel como culpable en el genocidio palestino. Algunos ejemplos de esta etiqueta son: *"Yanquis e israelís genocidas!", "Netanyahu, como candidato a dictador…", "Israel es el malo de la historia!".*

6. *Comentarios que señalan a la población palestina (Anti-palestino - Label: 2):* comentarios que señalan que los asesinatos a la población palestina son causados por la misma o la señalan de ser terroristas. Dentro de esta etiqueta se encuentran aquellos comentarios que señalan a la población palestina como culpable del conflicto, o atacan su religión y creencias además de su posición política. De la misma manera, se encuentran comentarios de tono agresivo hacia las personas de Gaza: *"Por qué los palestinos no hacen nada contra Hamas", "Ellos atacan y luego son las víctimas".*

7. Comentarios contra Hamás (Anti-Hamás - Label: 0): comentarios que explícitamente atacan a Hamás y lo señalan como grupo terrorista. Aquí se encuentran comentarios que señalan a Hamás como principal culpable del conflicto. *"Hamás debe de ser destruido"* o *"Terroristas los de Hamás"* son algunos ejemplos de señalamiento explícito del grupo armado.

El PLN se basa en anotaciones de sus textos para "evaluar las nuevas tecnologías del lenguaje humano y, fundamentalmente, para desarrollar modelos estadísticos fiables para la formación de estas tecnologías" (Ide y Pustejovsky, 2017, p. 2). Es por eso que el proceso de anotación se llevó a cabo de manera organizada y pensada en cuanto a la estructura lingüística de lo que se quiere analizar. Finlayson y Erjavec (2017) proponen un esquema como primer paso de anotación o etiquetaje de datos dentro de un proyecto que implique PLN. Los autores mencionan el esquema "MATTER", por sus siglas en inglés, como el más adecuado con algunas adecuaciones. Este esquema intenta establecer la ruta de la realización de una máquina de aprendizaje para PLN de la siguiente forma:

- M= modelado. Se establece el marco conceptual del proyecto.
- P= obtención (*procure*). Ver las anotaciones más apropiadas para el proceso. Paso agregado por los autores.
- A= anotación. La aplicación de anotaciones.
- TT= entrenamiento y testeo (*training and testing*). Entrenamiento y testeo del modelo de machine learning.
- E= evaluación.
- R= revisión.
- D= distribución. Compartir los resultados con la comunidad científica y el mundo en general. Paso agregado por los autores.

De esta manera, el proceso metodológico presente en este artículo cumple la estructura de "MATTER+PD". Primero se creó una estructura conceptual del modelo, ejemplificado aquí como una revisión de antecedentes y de contextualización del fenómeno a analizar. En segundo lugar, se estableció un proceso de generación de etiquetas o targets basados en la revisión de una cantidad estimada de comentarios y de análisis de contexto dispuestos por muestreo aleatorio. Se propuso como meta alcanzar 200 datos etiquetados por categoría para generar un total de 1400. Este total de datos etiquetados corresponden a un aproximado de 0,55% de los datos totales. La elección de la cantidad de comentarios se realizó teniendo en cuenta la potencia de la red neuronal a utilizar. Al ser BERT una red neuronal pre-entrenada en grandes cantidades de datos sin etiquetar, permite adquirir una comprensión profunda del lenguaje y las relaciones contextuales entre palabras. Así se realizó el modelo, la evaluación y la distribución. Este último paso de la estructura coincide con lo mencionado por Venturini y Munk (2022) en cuanto a un mapeo de controversias dispuesto para la discusión pública.

**Entendimiento del modelo**

El modelo BERT utilizado corresponde a una versión ajustada específicamente a las necesidades del estudio, utilizando para ello TensorFlow. Este ajuste se realizó con el fin de optimizar el desempeño del modelo en la tarea específica, considerando las características particulares de los datos y los objetivos del análisis. En este modelo primero "tokeniza" los datos, crea tuberías de entrada con TensorFlow junto con la API "tf.data" para su posterior entrenamiento y evaluación. Esta API se encuentra pre-entrenada y guardada como modelo "codificador para incrustaciones de texto con codificadores de transformador". Se espera un dict con tres en 32 Tensores como entrada: input_word_ids, input_mask, y input_type_ids" (Shrikant, 2023).

El tokenizador de BERT divide la cadena de texto en unidades básicas de significado o "tokens". Posteriormente, las cadenas pasan a una máscara de entrada: una matriz que se utiliza para indicar qué tokens son importantes para la tarea que se está realizando. A su vez los identificadores de tipo de entrada son números que se utilizan para representar el tipo de cada token. Luego de este proceso pasan a la capa BERT, la parte central del modelo. Esta capa es una red neuronal que aprende a representar el significado de las palabras en contexto. Al final todo se encapsula en Keras. Es precisamente la última capa de Keras asegura que el modelo funcione como clasificador. Al ser un modelo utilizado para otras actividades de clasificación, fue más fácil adecuar el código realizado en Python para que cumpliera con el entrenamiento y el criterio de predicción descritos por el investigador.

Dentro de BERT se encuentra un complejo de redes neuronales que utilizan aprendizaje en contexto. Una de las características clave de BERT es que es bidireccional. Esto significa que el modelo puede aprender el significado de una palabra en función de las palabras que la preceden y la siguen. Esto permite a BERT comprender mejor el contexto de una palabra y, por lo tanto, realizar mejor las tareas de PNL (Sanchez, 2020). Sin embargo, algunos artículos mencionan que puede presentar sesgos dependiendo de sus datos de entrenamiento. Así lo mencionan De Zárate et al. (2020) cuando revisa distintos modelos de *machine learning* para encontrar controversias. Los autores mencionan que, aunque no exista tal controversia, el modelo es capaz de percibirla gracias a su potencia.

Una vez dispuesto el modelo se procedió a entrenarle con los 1400 datos previamente etiquetados en 7 categorías. El modelo final fue entrenado en 15 épocas resultado de prueba y error. Cuando el entrenamiento culminó en la época número 15, el modelo presentó una precisión (accuracy) de entre 90% y 93%. Así, se tomó el modelo entrenado e inició el periodo de predicción para todos los datos. Después de realizar una limpieza de posibles comentarios previos al 7 de octubre del 2023, comentarios en blanco o con elementos que no consistían en una palabra o frase o emoji completo, se dejó un total de 253.925 comentarios con categoría predicha.

**Resultados**

Siguiendo las ideas expuestas en el apartado de revisión de literatura, es necesario iniciar con un contexto amplio que origine un marco de interpretación de los resultados obtenidos a partir del proceso de clasificación. Este contexto será realizado teniendo en cuenta información procedente de medios de comunicación similares a los elegidos en YouTube, de donde se obtuvieron los comentarios para este análisis de controversias. Así, el contexto estará enmarcado en un análisis de "agenda setting" que proponen McCombs y Valenzuela (2007) como la apropiación pública de las agendas de los medios de comunicación.

**Contexto de la controversia**

El ataque perpetrado por Hamás a Israel el 7 de octubre de 2023 dejó más de 900 personas muertas. Las primeras noticias que se publicaron después de este evento en medios informativos se enfocaban en qué es Hamás, cómo consiguió su poder, cuáles son sus principios y un recorrido del conflicto con Israel (Qué es Hamás, el grupo islamista militante que lanzó un ataque sin precedentes contra Israel, 2023). El ataque producido por Hamás fue calificado como "la operación más ambiciosa que Hamás ha lanzado desde Gaza y la penetración territorial más grave a la que Israel se ha enfrentado en una generación" (Qué es Hamás, el grupo islamista militante que lanzó un ataque sin precedentes contra Israel, 2023). La invasión de Hamás a Israel fue descrita como un suceso inédito donde más de "mil combatientes [atacan] en motocicletas, coches, planeadores motorizados y a pie" (Prange, 2023). Lo inusual de este ataque, comparado con otros, fue su grado de coordinación y la toma de rehenes israelíes.

La respuesta de Israel fue bombardear la Franja de Gaza, espacio territorial donde se concentraron los milicianos palestinos. Así, un día después del ataque de Hamás, se contabilizan al menos 413 personas muertas por parte de los ataques de Israel y más de 2.300 heridos (Guerra entre Israel y Gaza, 8 de octubre - Más de 1.100 muertos en el ataque de Hamás y la represalia de Israel, 2023). Las comunicaciones que se realizaron desde Israel apuntaban al anuncio de una guerra "larga y difícil" según el primer ministro israelí Benjamin Netanyahu, y algunas acciones inmediatas como el corte de la comunicación, electricidad, alimentos y combustibles a la Franja de Gaza. Lo cierto es que las condiciones en Gaza ya eran deplorables antes del anuncio de contraataque. Pocas horas de electricidad al día, carencia de agua potable y su clasificación de "territorio hostil" son condiciones que hacían a la Franja un lugar inhóspito (Prange, 2023).

La explicación contextual del conflicto Israel y Hamás juega un papel fundamental en los primeros días del mes de octubre de 2023. Estudiando las declaraciones del grupo, medios de comunicación señalan a Hamás como "organización palestina de islamista militante que niega el derecho de Israel a existir" (Cronología del conflicto entre Israel y Hamás, 2023). Aun así, es necesario mencionar que la postura palestina hacia Israel no es un monolito. Hamás no reconoce a Israel como Estado, mientras que Fatah, otro movimiento político que gobierna en Cisjordania y está de acuerdo con cumplir acuerdos internacionales relacionados al derecho de existencia del Estado israelí, le confronta y evidencia la división en la opinión pública palestina. Sin embargo, ambos movimientos abogan por un reconocimiento del pueblo palestino, así como un Estado Palestino (Barría, 2023). En los artículos de prensa consultados para este contexto no se mencionan las ocupaciones israelíes ni el contexto histórico de las razones por las que se creó el movimiento Hamás, que como se mencionó al inicio del artículo preceden por varias décadas el ataque del 7 de octubre. Los artículos, en cambio, describen los primeros enfrentamientos de la población palestina organizada y el ejército israelí hacia 1987, así como la prohibición de Hamás como grupo terrorista organizado en 1989 (Cronología del conflicto entre Israel y Hamás, 2023).

A un mes del recrudecimiento del conflicto entre Israel y Hamás, los reportes de noticieros internacionales empezaron a ser más amplios. Por un lado, se habla de las muertes de milicianos de Hamás por parte del Estado de Israel tras operaciones terrestres en Gaza. El Ministerio de Salud Palestino anotaba ya alrededor de 8.000 personas muertas, mucha de ellas civiles. Un punto que empieza a resonar es el número de niños muertos por ataques israelíes a la Franja. Israel, por su parte, insta a que la ciudadanía de Gaza se desplace hacia el sur del territorio para evitar un enfrentamiento directo. De igual manera, la Organización de Naciones Unidas (ONU) empieza a alertar de la necesidad de ayuda humanitaria y violaciones a derechos humanos a palestinos (Resumen de noticias de la guerra entre Israel y Hamas y la situación en Gaza del 29 de octubre de 2023, 2023). Así, empieza un reconocimiento del papel de Israel en el conflicto. El secretario general de la ONU, António Guterres, expresó en una reunión del Consejo de Seguridad que "nada puede justificar matar, herir y secuestrar deliberadamente a civiles, ni el lanzamiento de cohetes contra objetivos civiles" (El jefe de la ONU condena los ataques de Hamás, pero dice que "no ocurrieron de la nada" y embajador de Israel pide su dimisión, 2023).

En solo un mes, la guerra entre Israel y Hamás deja la misma cantidad de víctimas que el conflicto entre Rusia y Ucrania, que para este momento ya completaba los 20 meses. Aunque Netanyahu tenía el apoyo público para atacar a Gaza en Israel, a nivel internacional se encontraba presionado por las denuncias a violaciones de derechos humanos cometidos por el ejército de su país. Los medios de comunicación vaticinaban que no habría ningún acuerdo de tipo militar y que, en vez de buscar paz, "condenará a más guerras a las próximas generaciones de palestinos e israelíes" (Bowen, 2023). Sin embargo, a finales de noviembre empezaron a surgir ideas de treguas para el intercambio de rehenes entre Israel y Hamás. La noticia específica de que Hamás liberaría a "50 rehenes -todos ellos mujeres y niños" (Qué se sabe del acuerdo entre Israel y Hamás para declarar una tregua de 4 días e intercambiar rehenes, 2023), ocurre dentro del panorama dispuesto por Israel que incentiva a la liberación de rehenes. A su vez, Israel excarcela a 39 palestinos (Israel y Hamás intercambian prisioneros y rehenes durante la tregua de 4 días en Gaza, 2023). El intercambio de prisioneros es mediado por organismos internacionales como la Cruz Roja y la Medialuna Roja. Sin embargo, tras siete días de alto al fuego en total, Israel reanuda los ataques contra la Franja de Gaza (Fin de la tregua: se reanudan los combates entre Israel y Hamás y ya hay decenas de muertos en Gaza, 2023).

En este periodo los medios de comunicación empiezan a hablar de una posible salida al conflicto. Se señalan organizaciones internacionales, como la ONU, como idóneas para reconstruir los Estados. Se pone en la mesa la negociación la solución de dos Estados, que reconoce la legitimidad de que en el territorio existan dos entidades políticas: una israelí y otra palestina, como solución que implicaría que Israel cediera. Esta solución requeriría cesar todo intento de expansión de la frontera de Israel y pactaría demarcaciones para el Estado de Palestina. De igual forma, se puso en tela de juicio la continuidad de Netanyahu en el poder (Roiter, 2023). Este panorama se hizo más incierto cuando a finales de diciembre la cifra de muertos en Gaza aumentó hasta los 20.600, según información de autoridades palestinas. Hamás, por su parte, rechazó ceder el control de Gaza a cambio de un alto al fuego israelí. En respuesta, Netanyahu le comentó a su partido político en Israel: "no vamos a parar. Seguimos luchando, intensificaremos en los próximos días y los combates durarán mucho, no están cerca de concluir" (Peralta, 2023). De hecho, a finales de 2023, Israel advertía que la guerra continuaría durante todo el 2024, como efectivamente ocurrió. El 1 de enero de 2024 Israel bombardeó zonas de Gaza cobrando la vida de al menos 150 palestinos (Triviño, 2024).

Los primeros siete días del 2024, donde concluye el periodo de observación del presente artículo, señalaron la muerte del segundo al mando de Hamás, Saleh al-Arouri. Los medios advirtieron que esto podría llegar a desencadenar un recrudecimiento del conflicto (Esaá, 2024). A tres meses de inicio del conflicto, se comentó la victoria limitada de Israel, que no había logrado acabar con la presencia de Hamás en la Franja, y, a su vez, se señalaban las dudas para el futuro de la población palestina en caso de que Israel se hiciera con el control de toda Gaza. También se señaló a la ONU de ser incapaz de mantener la paz mundial, en vista de un posible recrudecimiento de los conflictos alrededor del mundo con Estados islamistas (Shbair, 2024). Hasta ese momento, los medios de comunicación reportaron más de 23.000 fallecidos y más de 58.000 heridos en Gaza. Las autoridades palestinas califican los actos de Israel como "masacre" y "genocidio" (Sube a más de 23.000 cifra de fallecidos en Gaza; ataque israelí mata a alto mando de Hezbolá en Líbano, 2024).

Este es el contexto en el que se enmarca la controversia, que permite analizar críticamente los resultados de clasificación a partir de inteligencia artificial, como se enuncia a continuación:

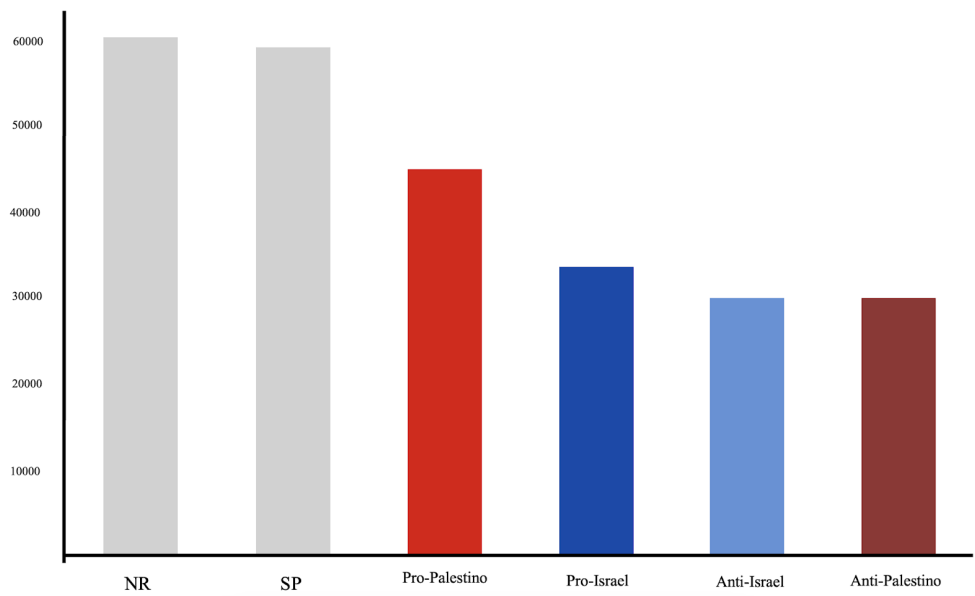

*Gráfica 1. cantidad de comentarios etiquetados por cada categoría.*

La gráfica 1 muestra la cantidad de comentarios etiquetados por cada una de las categorías. Organizadas de mayor a menor, no es de extrañar que los comentarios que más se presentan son aquellos que no tienen que ver con la controversia misma (Etiqueta 4, NR). Asimismo, los comentarios "sin postura" (Etiqueta 3, SP), que señalan tanto a Hamás como a Israel de culpables en la guerra, puntúan en la predicción de los comentarios. Esto concuerda con los resultados objetivos por Nushin et al. (2024) en Reddit.

Exceptuando las dos categorías que más puntúan en cantidad, le siguen las categorías Pro-Palestina (Etiqueta 6) y Pro-Israel (Etiqueta 5). Es aquí donde la controversia tiene lugar y se evidencian las dicotomías en los comentarios. Sin embargo, es notorio que la categoría Pro-Palestina tiene 34,16% más de comentarios que la categoría Pro-Israel. Por último, se encuentran las categorías Anti-Israel (Etiqueta 1) y Anti-Palestino (Etiqueta 2) con casi la misma cantidad de datos, 29.884 y 29.820 comentarios etiquetados respectivamente.

Es de destacar que no se encuentran etiquetas predichas de la categoría Anti-Hamás (Etiqueta 0) aunque en la fase de entrenamiento el modelo incluía 200 comentarios. Este comportamiento del modelo fue discutido y analizado previamente a finalizar la predicción para la totalidad de comentarios, al tratarse de la etiqueta más difícil de categorizar en el proceso de anotación. La revisión arrojó que los comentarios Anti-Hamás dentro de la base de datos de entrenamiento presentaban dificultades semánticas y contextos ambiguos que dificultan su predicción. Al analizar comentarios como *"Hay que matar a todos los terroristas de Gaza y no dejar a nadie"* se encontraba que, aunque explícitamente consistía en un comentario Anti-hamás, el significado intrínseco del mismo podría llevar a otras categorizaciones como Anti-palestino o Pro-Israel. Esta investigación no presenta una solución computacional para este aspecto, que se propone en cambio como una acción futura en el apartado de recomendaciones.

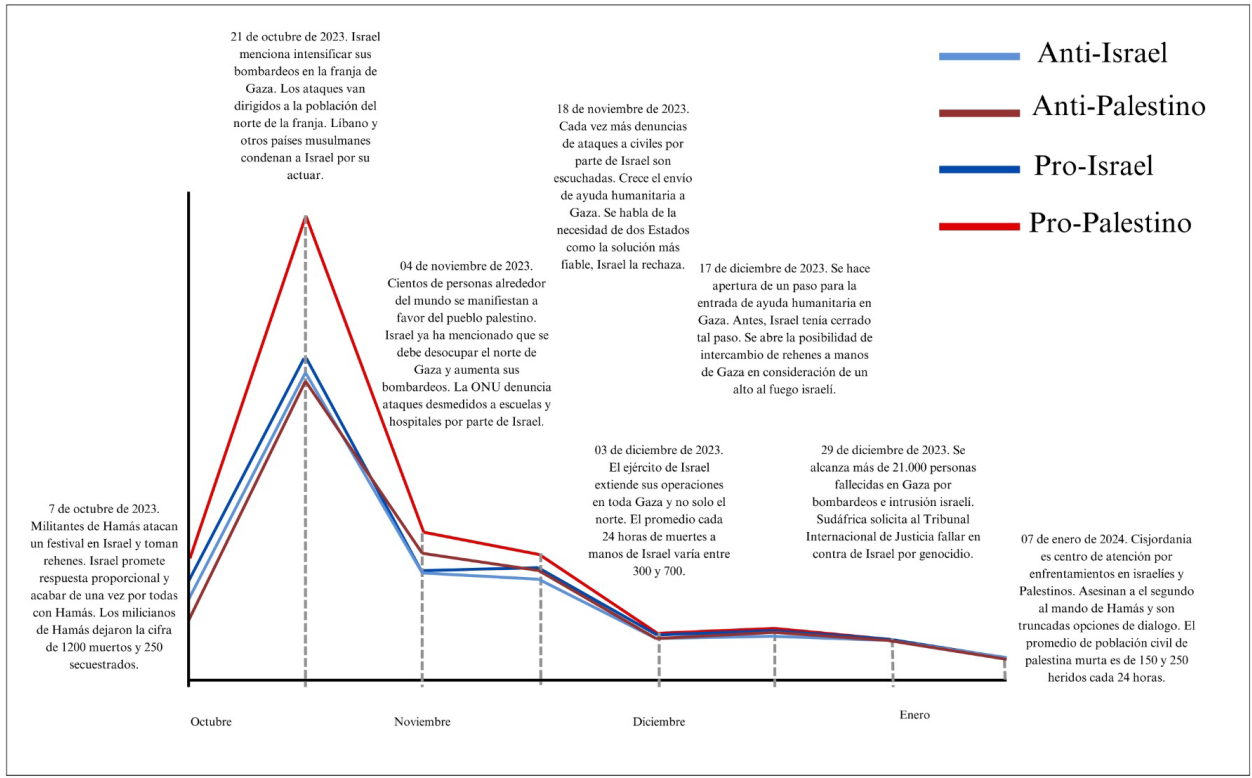

*Gráfica 2. Frecuencia de comentarios divididos cada dos semanas y su contexto.*

La gráfica 2 muestra, en corte de dos semanas, el comportamiento de las diferentes categorías clasificadas. El número de comentarios relacionados al conflicto Hamás-Israel se incrementó a más del doble en las dos primeras semanas posteriores al 7 de octubre. Es de destacar que el etiquetaje automático del modelo evidencia una presencia constante de comentarios Pro-Palestinos a lo largo del tiempo. Sin embargo, se destacan algunos puntos importantes en el comportamiento de la controversia en los tres meses analizados.

Los comentarios Anti-Palestinos no tuvieron presencia significativa, excepto en la última semana del mes de octubre y la primera semana del mes de noviembre. En este punto, la noticia de lo ocurrido era de interés global y se argumentaba comúnmente la legítima defensa de Israel. Como se analiza en el contexto de la controversia, varios medios de comunicación se enfocaban en el contexto histórico de la insurgencia del grupo palestino. En este periodo es cuando la categoría Anti-Palestino aparece en segundo lugar en cuanto a cantidad de comentarios presentes en los vídeos.

Hay que destacar también que el boom mediático que significó el ataque de Hamás, y luego el ataque Israel, dejó consigo más comentarios realizados en etapas tempranas de la temporalidad de la controversia analizada que hacia el final de la misma. Respondiendo a esta dinámica, el análisis temporal fue realizado según la fecha de realización del comentario y no de la publicación del vídeo. Así, se percibe que a pesar del paso del tiempo las personas que participan en la controversia no llegan a consensos y se percibe una fuerza estacionaria en su discurso.

Sin embargo, se da cuenta de un cambio en la composición de la controversia entre octubre y noviembre. Resulta interesante analizar que la cantidad de comentarios Anti-Israel desciende hacia la última posición y se mantiene así todo el mes de noviembre. En este periodo puntúan en segunda y tercera posición aquellos comentarios clasificados como Anti-Palestino y Pro-Israel. La dinámica del conflicto y su posición mediática hacen que la opinión pública tome una posición atacando a mercenarios palestinos y defendiendo el actuar israelí.

Retomando la información empírica que se dio cuenta con anterioridad, se puede concluir que la controversia Hamás-Israel es un ejemplo del análisis de "agenda setting" propuesto por McCombs y Valenzuela (2007). Al inicio de la controversia, los medios de comunicación fijaron la agenda informativa a partir de lo que se conocía de la guerra de Israel frente al ataque de Hamás. McCombs y Valenzuela (2007) afirman: "cuanto mayor es nuestra necesidad de orientación, más tendemos a buscar información, depender de los medios de comunicación y estar predispuestos a los efectos de fijación de la agenda" (p. 46). De esta manera se explica el por qué la cantidad de comentarios al inicio de la revisión de la controversia supera en cantidad a las otras semanas analizadas. La falta de claridad sobre el conflicto hizo que la búsqueda por información fuese cada vez mayor, puesto que "cuanto mayor es la incertidumbre sobre un tema, mayor será [entonces] la necesidad de orientación" (McCombs y Valenzuela, 2007, p. 46).

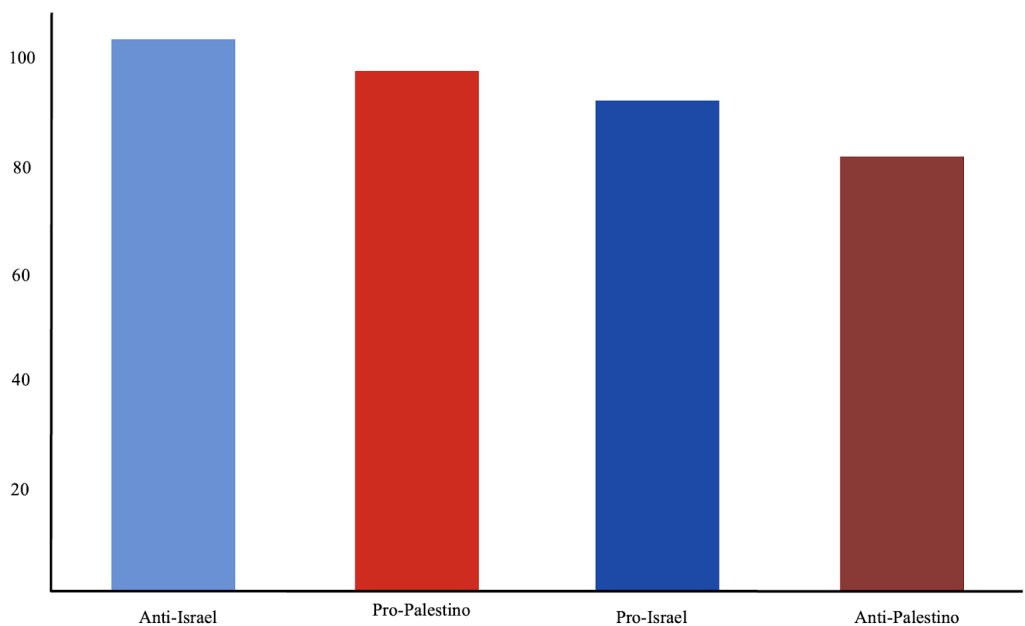

*Gráfico 3. Cantidad de comentarios para el mes de enero de 2024 dividido por categorías.*

Sin embargo, en la gráfica 3 se puede observar un ligero cambio hacia el final de las líneas. La gráfica 3 muestra que, pese a que fueron tomados comentarios realizados solo en los primeros ocho días del mes de enero, el modelo identifica más comentarios en la categoría Anti-Israel. Este cambio de comportamiento en la controversia puede significar un cambio en la opinión pública en general. Luego de más de 100 días del recrudecimiento del conflicto, las acciones realizadas por Israel en Gaza y la crisis humanitaria que vive el pueblo palestino aparecen más en el cubrimiento de los medios de comunicación.

Lo analizado en la contextualización permite inferir que las crecientes cifras de fallecidos en Gaza y las denuncias realizadas por organismos internacionales como la ONU generaron un cambio en la opinión pública reflejada en los comentarios. Así, con el paso del tiempo se da entrada a nuevos actores políticos y sociales a la controversia. Lo anterior demuestra que la controversia cumple con la característica de imposibilidad de estaticidad, en este caso, permeada por la opinión pública (Venturini y Munk, 2022).

La modificación del discurso ante los crímenes de lesa humanidad perpetrados por Israel hacia mitad y finales de la temporalidad de la controversia analizada muestra cómo la agenda de los medios de comunicación puede ocasionar un cambio en la controversia misma. McCombs y Valenzuela (2007) mencionan que "el papel de los medios de comunicación no se limita a centrar la atención del público en un conjunto particular de cuestiones, sino que también influye en nuestra comprensión y perspectiva sobre los temas de las noticias" (p. 47). Así, vemos cómo la agenda mediática puede influir en la disposición de una controversia ocasionando, en este caso, que la misma cambie en clasificación y pase de una preponderancia Pro-Palestina a Anti-Israel.

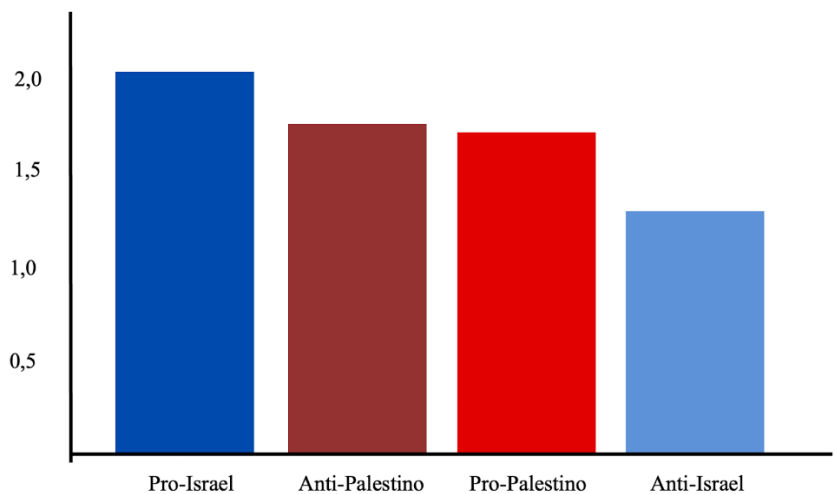

*Gráfico 4. Promedio de "likes" por cada categoría.*

Otro punto importante a analizar es el promedio de "likes" por categoría. Exceptuando las categorías NR y SP, en promedio los comentarios con más "likes" hacen parte de las categorías Pro-Israel y Anti-palestino. Esto quiere decir que, a pesar de que existen más comentarios Pro-Palestino, los comentarios más populares son aquellos que defienden a Israel o atacan al pueblo palestino. Se concluye así que, aunque la participación de comentarios Pro-israelíes y Anti-palestinos son menos frecuentes que los comentarios Pro-Palestinos y Anti-Israelí, son los que más reciben apoyo por parte de los usuarios de YouTube.

**Conclusión**

El PLN es percibido dentro de este artículo como una herramienta central para entender la controversia Hamás-Israel. El proceso metodológico utilizado demuestra que el uso de herramientas computacionales para el entendimiento de asuntos sociales potencializa su comprensión, siempre y cuando se apliquen pautas reflexivas y contextuales para su utilización. Así, dentro de esta investigación se hizo uso de inteligencia artificial -más específicamente se realizó el entrenamiento de un modelo de predicción de clasificación BERT con 1.400 comentarios, obtenidos principalmente de páginas de medios de comunicación en YouTube- para que adelantara una clasificación automática de 253.925 comentarios en total. A partir del proceso de etiquetado, este proyecto analiza las diferentes orientaciones de los comentarios a lo largo del tiempo en términos de su postura en el conflicto.

Algunos de los principales hallazgos encontrados en el mapeo de la controversia Hamás-Israel a finales de 2023 e inicios de 2024 se centran en el impacto de la noticia y la consolidación de la controversia misma a lo largo del tiempo. Se revela, además, una tensión entre la afinidad marcada por "likes" y la cantidad de comentarios. Aunque hubo mayor cantidad de comentarios Pro-Palestino, los comentarios que en promedio recibieron más "me gusta" son aquellos de categoría Pro-Israel. Igualmente, entre los hallazgos más importantes del análisis de la controversia se encuentra la mutación de la frecuencia de comentarios y de la controversia misma. Durante los primeros tres meses del recrudecimiento del conflicto, esta pasó de una mayor frecuencia de comentarios Pro-Palestino hacia el inicio de la controversia a una mayor cantidad de comentarios etiquetados como Anti-Israel.

La controversia Hamás-Israel ejemplifica el modelo de *agenda setting* propuesto por McCombs y Valenzuela (2007) al mostrar cómo los medios de comunicación influyen en la percepción pública y en la búsqueda de información en situaciones de incertidumbre. Al inicio del conflicto, la cobertura mediática se centró en la respuesta de Israel al ataque de Hamás, lo que generó un aumento significativo en la atención del público (dada su necesidad de información y falta de claridad del conflicto) que derivó en una mayor cantidad de comentarios. Conforme avanzó la controversia, el enfoque mediático se desplazó hacia la denuncia de crímenes de lesa humanidad cometidos por Israel, lo que provocó un cambio en la percepción pública, pasando de una postura Pro-Palestina a una visión Anti-Israel. Este análisis evidencia cómo los medios no sólo pueden contribuir a fijar la agenda informativa, sino que también influyen en la comprensión y las actitudes del público hacia los acontecimientos.

Aunque exploratoria, la presente investigación marca puntos importantes para el desarrollo de modelos de aprendizaje de máquina que pueden ayudar a investigadores de diversas disciplinas a identificar y analizar elementos concluyentes para la comprensión de fenómenos sociales. Los puntos claves de esta investigación demuestran la posibilidad de realizar generalizaciones sobre una controversia presente en la opinión pública a partir de procesos computacionales. El intento de mediación entre el entendimiento de un mapeo de controversias desde la comunicación y ciencias sociales, y aquel rico en procedimientos computacionales, da como resultado la compresión de la opinión pública respecto a los sucesos acontecidos en los primeros tres meses del conflicto Hamás-Israel vigente a inicios de 2024. Esta exploración se enriquece con la interacción entre diferentes disciplinas con el fin realizar análisis más robustos y críticos basados en la interdisciplinariedad.

**Consideraciones y recomendaciones**

Para finalizar este artículo es menester realizar algunas recomendaciones futuras en las que se enlistan:

1. Entrenar un modelo con más datos que le permitan a la máquina identificar nuevas categorías existentes no presentes en este documento. La eficiencia del modelo puede ser mejorada revisando la disposición del mismo.
2. Analizar la etiqueta "Sin Postura" (SP) pues, aunque el objetivo de la investigación eran las posturas polares en la controversia, esta etiqueta puede tomarse como postura en sí misma y ser analizada.
3. Se deben analizar y reconsiderar las etiquetas teniendo en cuenta razones implícitas de escritura de los comentarios, haciendo un análisis reflexivo de las posturas tomadas dentro de la controversia. Esta tarea se debe entender como una investigación científico-social y crítica que muta con la controversia a lo largo del tiempo.
4. Continuar con la comprensión del mapeo de controversia presente otros idiomas para obtener nuevas perspectivas sobre el tema estudiado.